\documentclass[conference]{IEEEtran}
\usepackage{cite}
\usepackage{amsmath,amssymb,amsfonts}
\usepackage[noend]{algorithmic}
\usepackage{graphicx}
\usepackage{textcomp}
\usepackage{xcolor}
\def\BibTeX{{\rm B\kern-.05em{\sc i\kern-.025em b}\kern-.08em
    T\kern-.1667em\lower.7ex\hbox{E}\kern-.125emX}}

\usepackage{hyperref}
\usepackage{algorithm}
\usepackage {booktabs}
\usepackage{soul}

\newcommand\figref[1]{\textbf{Fig.~\ref{#1}}}

\newcommand\algref[1]{\textbf{Algorithm~\ref{#1}}}
\newcommand\secref[1]{\textbf{Section~\ref{#1}}}

\makeatletter
\def\ps@IEEEtitlepagestyle{
    \def\@oddfoot{\mycopyrightnotice}
}
\def\mycopyrightnotice{\begin{minipage}{\textwidth}
\centering \scriptsize {\color{blue} \copyright~2022 IEEE. Personal use of this material is permitted. Permission from IEEE must be obtained for all other uses, in any current or future media, including reprinting/republishing this material for advertising or promotional purposes, creating new collective works, for resale or redistribution to servers or lists, or reuse of any copyrighted component of this work in other works.}
\end{minipage}}
\makeatother

\begin{document}
\title{Knowledge-Driven Program Synthesis via Adaptive Replacement Mutation and Auto-constructed Subprogram Archives}
\author{\IEEEauthorblockN{Yifan He}
\IEEEauthorblockA{\textit{University of Tsukuba}\\
Tsukuba, Japan \\
he.yifan.xs@alumni.tsukuba.ac.jp}
\and
\IEEEauthorblockN{Claus Aranha}
\IEEEauthorblockA{\textit{University of Tsukuba}\\
Tsukuba, Japan \\
caranha@cs.tsukuba.ac.jp}
\and
\IEEEauthorblockN{Tetsuya Sakurai}
\IEEEauthorblockA{\textit{University of Tsukuba}\\
Tsukuba, Japan \\
sakurai@cs.tsukuba.ac.jp}
}
\maketitle

\begin{abstract}
We introduce Knowledge-Driven Program Synthesis (KDPS) as a variant of the program synthesis task that requires the agent to solve a sequence of program synthesis problems. In KDPS, the agent should use knowledge from the earlier problems to solve the later ones. We propose a novel method based on PushGP to solve the KDPS problem, which takes subprograms as knowledge. The proposed method extracts subprograms from the solution of previously solved problems by the \emph{Even Partitioning} (EP) method and uses these subprograms to solve the upcoming programming task using \emph{Adaptive Replacement Mutation} (ARM). We call this method PushGP+EP+ARM. With PushGP+EP+ARM, no human effort is required in the knowledge extraction and utilization processes. We compare the proposed method with PushGP, as well as a method using subprograms manually extracted by a human. Our PushGP+EP+ARM achieves better train error, success count, and faster convergence than PushGP. Additionally, we demonstrate the superiority of PushGP+EP+ARM when consecutively solving a sequence of six program synthesis problems.
\end{abstract}


\section{Introduction}
\label{sec:intro}
\emph{Program Synthesis} (PS) are techniques that automatically compose computer programs to solve a certain task. PS is useful in fields such as automatic bug fixing, automatic program completion, and low-level code development. PS is a key issue in Artificial General Intelligence~\cite{o2020automatic}. \emph{Genetic Programming} (GP)~\cite{koza1992programming} is an Evolutionary Algorithm that searches computer programs by selecting and updating a population of program candidates. Some GP variants~\cite{helmuth2018program,kelly2019improving,helmuth2020explaining} can solve problems in a famous PS benchmark suite~\cite{helmuth2015general} efficiently.

However, the difference between GP and a human programmer is still obvious. As an Evolutionary Algorithm, GP heavily utilizes random sampling; while a human programmer does not write random programs. Humans write programs based on their knowledge, either the domain knowledge about the problem or the programming skills from previous experiences.

Recently, several studies~\cite{hemberg2019domain,sobania2019teaching,helmuth2020genetic,wick2021getting,yifan2022incorporating} have attempted to incorporate knowledge in PS, improving the synthesis performance. However, some of these methods have drawbacks in requiring extra information~\cite{hemberg2019domain,sobania2019teaching} and human efforts~\cite{yifan2022incorporating}.

In our prior study, we proposed the \emph{Adaptive Replacement Mutation} (ARM)~\cite{yifan2022incorporating}. The ARM is a mutation method designed for a well-known GP variant called PushGP~\cite{helmuth2018program}. ARM uses subprograms from an archive as knowledge, automatically selecting useful subprograms from the archive according to the search history. Although the ARM provides a way to use existing knowledge from an archive, the archive itself was made by a human. Moreover, it is questionable whether the subprograms written by humans are included in the programs generated by PushGP.

In this study, we focus on the task where an agent is required to solve a sequence of PS problems. The agent should learn knowledge from each problem in the sequence and apply this knowledge to improve its performance in later problems. Ideally, this procedure should not require human interference or extra external information. We call this task the \textbf{\emph{Knowledge-Driven Program Synthesis}} (KDPS) problem (\secref{sec:prob}).


We propose a novel method to solve the KDPS problem based on PushGP~\cite{helmuth2018program}. This method takes subprograms as knowledge. The proposed method consecutively solves programming tasks, extracts subprograms from the solution of solved problems, and uses subprograms to solve an upcoming problem. To extract subprograms, we propose \emph{Even Partitioning} (EP) which divides a solution into several parts with equal lengths. To use these subprograms, we apply ARM~\cite{yifan2022incorporating}. We name our method PushGP+EP+ARM. The details of the proposed methods, including EP and ARM, are given in~\secref{sec:method}.


We analyze the proposed method in two KDPS tasks. The ``composite task'' (\secref{sec:experiment-i}) includes three ``composite'' PS problems. For each problem, PushGP+EP+ARM prepares the knowledge archive based on the component problems. The ``sequential task'' (\secref{sec:experiment-ii}) has six problems that must be solved in sequence, including the composite and component problems of the previous ``composite task''. PushGP+EP+ARM updates the knowledge archive at each step of the sequence. PushGP+EP+ARM achieves a better overall success rate and convergence speed in the composite task, and in the later stages of the sequential task, showing that it can create a useful knowledge archive. However, the comparison with a human-curated archive shows that there is still room for improvement.

Our main contributions are as follows.
\begin{enumerate}
    \item We introduce a new type of task called the KDPS problem. KDPS includes a sequence of single PS problems. The agent is required to solve the single PS problems, extract knowledge, and use it in the later PS problems.
    \item We propose EP to extract subprograms from the solution of a solved problem. We propose PushGP+EP+ARM to solve KDPS problems based on EP and our previous work on ARM~\cite{yifan2022incorporating}.
    \item We provide our implementation of the proposed method based on PyshGP~\cite{pantridge2017pyshgp} and experimental scripts in an online repository~\footnote{\href{https://github.com/Y1fanHE/ssci2022}{https://github.com/Y1fanHE/ssci2022}}
\end{enumerate}

\section{Background}
\label{sec:back}
\subsection{Program Synthesis}
PS, also known as automatic programming, focuses on building an intelligent agent that writes computer programs to solve specific tasks with minimal human effort. Programs are sequences of instructions and a task could be described as a set of I/O examples. For instance, a task that adds two integers could be describes as \texttt{\{[in=(1,1), out=2], [in=(1,2), out=3], ...\}}.

Therefore, PS can be formalized as the optimization problem in~\eqref{eq:ps}, which searches a sequence of instructions to minimize the difference between actual and expected program outputs. In~\eqref{eq:ps}, $\mathbf{p}$ is a program while $P_{instr}$ includes all feasible programs based on the instruction set $instr$; $in_{i}$ and $out_{i}$ are the $i$-th I/O example.

\begin{equation}
\label{eq:ps}
\min_{\mathbf{p}\in{P_{instr}}}\Sigma_{i=1}^{N}||\mathbf{p}(in_{i})-out_{i}||
\end{equation}

\subsection{PushGP}
Since Koza's first work on his tree-based GP~\cite{koza1992programming}, several variants of GP~\cite{o2001grammatical,spector2001autoconstructive,forstenlechner2018extending,helmuth2018program} have been applied to solve PS problems. Among these methods, we highlight PushGP~\cite{spector2001autoconstructive,helmuth2018program,helmuth2020explaining}, which generates programs based on a Turing-complete language called Push.

Push uses the list to store a program and runs it using multiple stacks of different data types based on the following rules.

\begin{enumerate}
    \item To execute an instruction, the interpreter pops the required arguments from the corresponding stacks.
    \item After executed the instruction, the results are pushed to the corresponding stacks.
    \item If the interpreter cannot find enough arguments from the stacks, the instruction will be skipped.
\end{enumerate}

Based on the third rule, any sequence of Push instructions can form a valid Push program. Helmuth et al.~proposed a variant of PushGP using lexicase selection and uniform mutation by addition and deletion (UMAD)~\cite{helmuth2018program}. This variant was further improved by applying down-sampled lexicase selection~\cite{helmuth2020explaining}.

\subsection{Incorporating knowledge in Program Synthesis}
An obvious difference between GP and human programmers is that a human can learn knowledge from his experiences and use knowledge in upcoming tasks. Several recent works have attempted to incorporate knowledge in PS~\cite{sobania2019teaching,hemberg2019domain,helmuth2020genetic,wick2021getting,yifan2022incorporating,lehman2022evolution}. Some studies require extra information such as text description~\cite{hemberg2019domain} and human-written code~\cite{sobania2019teaching,yifan2022incorporating,lehman2022evolution}. 

Helmuth et al. proposed to transfer the instructions from the solution of other problems to construct the instruction set of PushGP~\cite{helmuth2020genetic}. Wick et al. proposed to use the whole individuals of a problem as a part of the initial population when solving similar problems~\cite{wick2021getting}. Both studies~\cite{helmuth2020genetic,wick2021getting} extract knowledge from the solutions of the problems that have been solved by GP.

Recently, we came up with a mutation method that allows using subprograms in a prepared archive with PushGP~\cite{yifan2022incorporating}. Compared with the study by Helmuth~\cite{helmuth2020genetic}, a subprogram can capture more information from the solution than a single instruction. Compared to Wick's study~\cite{wick2021getting}, the way to use an external archive of knowledge might be more efficient when the number of the problems to transfer is large. However, in our prior study~\cite{yifan2022incorporating}, the subprogram archives are extracted by hand from human-written solutions. This step is non-trivial and requires a lot of human effort. Also, considering the case where a GP is able to extract knowledge from programs it wrote in the past, it is questionable whether the source programs would look similar to human written ones.


\section{Knowledge-Driven Program Synthesis}
\label{sec:prob}
Human programmers can learn from the programming problems that they have solved before and apply what they have learned to solve the upcoming problems. We suggest that the ability of learning is vital for generating complex programs.

Therefore, we introduce a type of task where an agent is required to solve a sequence of PS problems. Usually, these PS problems are related to each other. When solving a new problem, the agent should use the knowledge learned from previously solved problems. We call this type of task \emph{Knowledge-Driven Program Synthesis} (KDPS).

The formalization of KDPS is presented in~\eqref{eq:kdps}. $S^j$ is one of the $M$ PS problems in~\eqref{eq:ps} to solve, containing $N_j$ pairs of I/O. $\hat{\mathbf{p}}^j$ and $K^j$ are the solution program and the knowledge that learned from solving $S^j$, respectively. $S^j$ is solved by $\mathrm{Solve}(\cdot)$ based on the knowledge from the previously solved problems $K^{j-1}\cup\dots{K}^0$. $K^0$ is the initial knowledge before solving the first problem $S^1$. By default, $K^0$ is empty.

\begin{equation}
\label{eq:kdps}
\begin{cases}
     & \hat{\mathbf{p}}^j,K^j\leftarrow \mathrm{Solve}(S^j|K^{j-1}\cup\dots\cup{K}^0) \\
s.t. & S=\{S^1,\dots,S^M\} \\
     & S^j=\{({in}^j_1,{out}^j_1),\dots,({in}^j_{N_j},{out}^j_{N_j})\} \\
     & K^0=\emptyset
\end{cases}
\end{equation}

To learn knowledge from solving a problem, it is not necessary to use its solution program. However, as an initial step, we assume all the knowledge from solving $S^j$ are extracted from its solution programs $\hat{\mathbf{p}}^j$ as in~\eqref{eq:extract}.

\begin{equation}
\label{eq:extract}
K^j\leftarrow \mathrm{Extract}(\hat{\mathbf{p}}^j)
\end{equation}

\section{Proposed Method}
\label{sec:method}
\begin{figure}
\centering
\includegraphics[width=0.8\columnwidth]{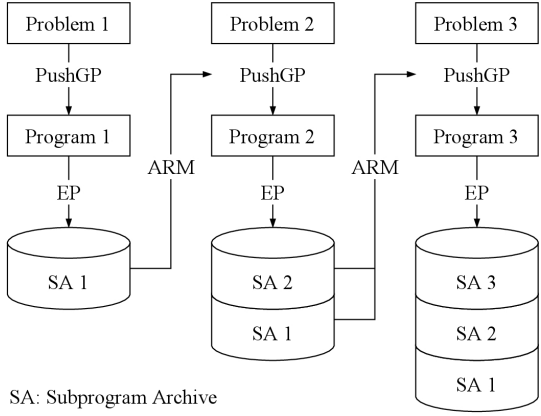}
\caption{An example of solving Knowledge-Driven Program Synthesis with the proposed method. The method solves problems with PushGP, extract subprograms with \emph{Even Partitioning} (EP), and uses subprograms with \emph{Adaptive Replacement Mutation} (ARM).}
\label{fig:kdps}
\end{figure}

To solve KDPS problems, we propose a method based on PushGP~\cite{helmuth2018program} to consecutively solve programming tasks, extract knowledge from the solutions, and utilize knowledge in the next problem. This method is entirely automated.

We use subprograms, the sub-sequences of Push instructions, as knowledge. The subprograms hold partial information about the original program. Moreover, any sequence of Push instructions is valid to run. Therefore, we can easily take a subprogram and use it in a different program.

\figref{fig:kdps} illustrates an example of solving a KDPS problem (a sequence of three PS problems) with our proposed method. Our method solves Problem 1 with PushGP~\cite{helmuth2018program} and extracts Subprogram Archive (SA) 1 from its solution (Program 1) with \emph{Even Partitioning} (EP). This archive is used by \emph{Adaptive Replacement Mutation} (ARM)~\cite{yifan2022incorporating} when searching for the solution of Problem 2. Similarly, we use SA 1 and SA 2 when solving Problem 3. The details of EP and ARM are provided in the next two subsections.

\subsection{Even Partitioning}
EP is a simple method that divides a program into $n$ parts with equal lengths. For example, a program with 15 instructions is divided into subprograms with lengths of (3, 3, 3, 3, 3) if $n$ = 5; the same program is divided into subprograms with lengths of (4, 4, 4, 3) if $n$ = 4.

Before dividing the solution program, a simplification operation is performed to remove the redundant instructions (i.e., instructions without enough arguments to execute). After the dividing step, the subprograms are stored into an archive for the future use.

\subsection{Adaptive Replacement Mutation}
\begin{algorithm}
\caption{PushGP with Adaptive Replacement Mutation}
\begin{algorithmic}[1]
\REQUIRE subprogram archive $K$ where every subprogram $\mathbf{k}$ holds a quality metric $Q_\mathbf{k}$ = 0;
\STATE $P$ \textleftarrow\ initialize\_population();
\WHILE{termination criteria is not satisfied}
    \STATE $P'$ \textleftarrow\ $\emptyset$;
    \FOR{$i$ \textleftarrow\ 1 to $|P|$}
        \STATE $\mathbf{p}$ \textleftarrow\ lexicase\_selection($P$);
        \IF{rand() $<$ $r_{arm}$}
            \IF{rand() $<$ $r_{prop}$}
                \STATE $\mathbf{k}$ \textleftarrow\ proportional\_selection($K$);
            \ELSE
                \STATE $\mathbf{k}$ \textleftarrow\ random\_selection($K$);
            \ENDIF
            \STATE $\mathbf{p}'$ \textleftarrow\ replacement\_mutation($\mathbf{p}$, $\mathbf{k}$);
            \IF{$f(\mathbf{p}') \prec f(\mathbf{p})$}
                \STATE $Q_\mathbf{k}$ \textleftarrow\ $Q_\mathbf{k}$ + 1;
            \ENDIF
        \ELSE
            \STATE $\mathbf{p}'$ \textleftarrow\ umad\_mutation($\mathbf{p}$);
        \ENDIF
        \STATE $P'$ \textleftarrow\ $P'$ $\cup$ \{$\mathbf{p}'$\};
    \ENDFOR
    \STATE $P$ \textleftarrow\ $P'$;
\ENDWHILE
\end{algorithmic}
\label{alg:arm}
\end{algorithm}
ARM~\cite{yifan2022incorporating} is a mutation method proposed for PushGP, which incorporates a prepared archive of subprograms during the search. ARM contains a Replacement Mutation (RM) and an adaptive strategy.

RM requires a parent candidate (of length $l_1$) from the PushGP population and a subprogram (of length $l_2$) from the prepared archive. RM replaces a random partition (of length $l_2$) of the parent candidate using the subprogram. If $l_1<l_2$, the entire parent is replaced by the subprogram.

The adaptive strategy in ARM is designed to automatically select helpful subprograms when an archive contains both helpful and unhelpful subprograms. The idea is similar to parameter adaptation in many Self-adaptive Evolutionary Algorithms such as JADE~\cite{zhang2009jade}. Originally, ARM collects and stores the helpful subprograms that improve the parent candidates by RM into a working archive. This working archive, which consists of helpful subprograms, is used as one of the sources to select subprograms in the later generations.

In this study, we use a slightly different implementation of ARM from the original~\cite{yifan2022incorporating}, using proportional selection to select subprograms rather than collecting subprograms in an extra archive. The proportion to select a subprogram depends on how many times that it improves parents.

We provide the pseudo code of PushGP (using lexicase selection and UMAD mutation) with ARM in~\algref{alg:arm}. $Q_\mathbf{k}$ is the count that a subprogram $\mathbf{k}$ improves the parents during the search. The probability in proportional selection is computed as in~\eqref{eq:prop}.

\begin{equation}
\label{eq:prop}
p(\mathbf{k})=\frac{Q_\mathbf{k}}{\Sigma_{\mathbf{k}_i\in{K}}Q_{\mathbf{k}_i}}
\end{equation}

$r_{arm}$ is the probability to perform ARM and $r_{prop}$ is the probability to perform the proportional selection of subprograms. In Line 12 of~\algref{alg:arm}, the symbol ``$\prec$'' means ``better than''. In our implementation, a solution is better than another if it solves more I/O cases (i.e., contains more ``0'' in its error vector).

In some cases, the subprograms may include more inputs than the current problem (e.g., a subprogram contains \texttt{input\_3} while the current problem only takes two inputs). We replace the input in the subprograms with a random input of the current problem.

\section{Experiment I: Composite KDPS Task}
\label{sec:experiment-i}
In this experiment, we focus on an intermediate step of the KDPS problem (\figref{fig:kdps}). That is, to solve a composite problem with our proposed method after solving its sub-problems.

\subsection{Comparison methods}

We compare the following three methods.


\begin{itemize}
\item \textbf{PushGP+EP+ARM}: PushGP with ARM; the subprogram archives are extracted using EP.
\item \textbf{PushGP+HP+ARM}: PushGP with ARM; the subprogram archives are extracted by human (HP: human partitioning).
\item \textbf{PushGP}: the original PushGP~\cite{helmuth2018program}.
\end{itemize}

\subsection{Experimental procedures}

We use PushGP~\cite{helmuth2018program} to solve three problems in PSB1~\cite{helmuth2015general}. They are ``small or large'' (SL), ``compare string lengths'' (CSL), and ``median'' (MD). We take the best and shortest solutions among 25 runs (after 5000 steps of simplification) of the three problems to generate subprogram archives. For PushGP+EP+ARM, We use EP to get five equal-length subprograms for every best and shortest solution automatically. The subprograms used by PushGP+HP+ARM are partitions of the same solutions, however, devised by a human.

We then solve the composite problems of SL, CSL, and MD. When solving a composite problem, PushGP+EP+ARM and PushGP+HP+ARM will use subprogram archives generated from solutions of the corresponded sub-problems by EP and HP, respectively. We compare the three methods on three composite problems.

\begin{itemize}
\item \textbf{Median String Length (MDSLEN)}: given 3 strings, print the median of their lengths.
\item \textbf{Small or Large Median (SLMD)}: given 4 integers $a$, $b$, $c$, $d$, print ``small'' if median($a$,$b$,$c$) $<$ $d$ and ``large'' if median($a$,$b$,$c$) $>$ $d$ (and nothing if median($a$,$b$,$c$) = $d$).
\item \textbf{Small or Large String (SLSTR)}: given a string $n$, print ``small'' if len($n$) $<$ 100 and ``large'' if len($n$) $\geq$ 200 (and nothing if 100 $\leq$ len($n$) $<$ 200).
\end{itemize}

MDSLEN is the composite problem of MD and CSL; SLMD is composed from SL and MD; SLSTR is a composite of SL and CSL.

\subsection{Parameter settings}
For all three methods, we use a population size of 1000 and a maximum generation of 300. The UMAD mutation in all three methods is set with addition rate of 0.09 and deletion rate of 0.0826 based on Helmuth's study~\cite{helmuth2018program}. For PushGP+EP+ARM and PushGP+HP+ARM, the rate to perform ARM $r_{arm}$ is 0.1 and the rate to perform the proportional selection of subprograms $r_{prop}$ is 0.5 based on the original paper of ARM~\cite{yifan2022incorporating}. For every algorithm, we run 25 repetitions on every problem.

\subsection{Experimental results}
\begin{figure*}
\centering
\includegraphics[width=\textwidth]{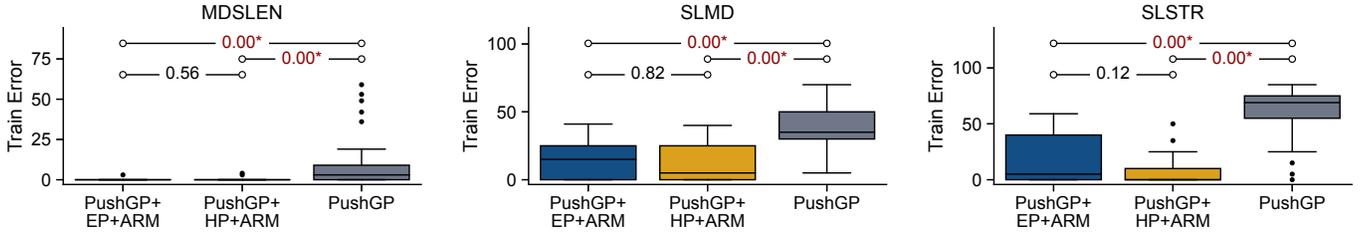}
\caption{\textbf{Experiment I}: Train error in 25 runs. The value on a line segment is the p-value of Wilcoxon rank sum test between two groups. The p-value is marked with an asterisk and red color if the difference between two groups is significant. PushGP+ARM+EP holds a significantly lower train error than PushGP and a comparable train error compared to PushGP+ARM+HP.}
\label{fig:train}
\end{figure*}
\begin{figure*}
\centering
\includegraphics[width=\textwidth]{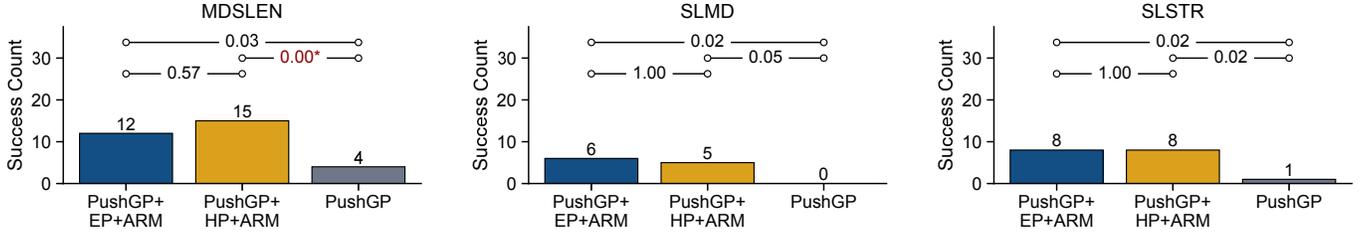}
\caption{\textbf{Experiment I}: Test success count in 25 runs. The number above a bar is the success count. The value on a line segment is the p-value of Fisher's exact test between two groups. The p-value is marked with an asterisk and red color if the difference between two groups is significant. PushGP+ARM+EP holds higher success counts than PushGP but without statistical significance.}
\label{fig:count}
\end{figure*}
\begin{figure*}
\centering
\includegraphics[width=\textwidth]{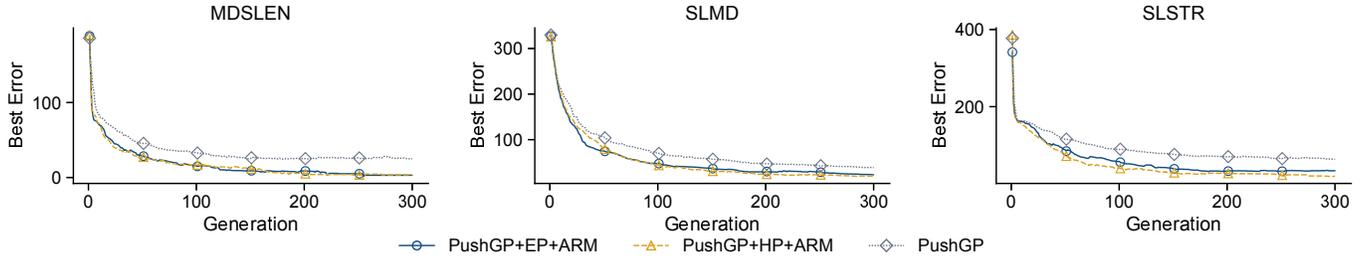}
\caption{\textbf{Experiment I}: Average of the best train error in the population by generations. PushGP+ARM+EP converges much faster than PushGP but slightly slower (SLSTR) or at the similar speed (MDSLEN and SLMD) compared to PushGP+ARM+HP.}
\label{fig:error}
\end{figure*}
\figref{fig:train} presents the error in the training period of the three methods in 25 runs. The value on a line segment is the p-value of the Wilcoxon rank sum test between two groups. The p-value is marked with an asterisk and red color if the difference between two groups is significant with a 95\% family confidence level (i.e., the individual confidence level is computed by \v{S}id\'ak correction\footnote[2]{The individual confidence levels in~\secref{sec:experiment-i} and~\secref{sec:experiment-ii} are 99.4\% and 99.1\%, respectively.}). The proposed PushGP+EP+ARM outperforms the original PushGP with a significant difference in the training error. However, compared with PushGP+HP+ARM, the difference is not statistically significant.

\figref{fig:count} shows the success counts in the test period of the three comparison methods in 25 runs. We count a run as a successful run only when it passes all I/O cases in both training and testing data. In~\figref{fig:count}, the number above a bar is the success count; the value on a line segment is the p-value of Fisher’s exact test between two groups. The p-value is marked with an asterisk and red color if the difference between two groups is significant with a 95\% family confidence level (i.e., the individual confidence level is computed by \v{S}id\'ak correction\footnotemark[2]). Compared to PushGP, our PushGP+EP+ARM achieves higher success counts, however, without statistical significance. Compared to the method using human-made subprograms (PushGP+HP+ARM), our proposed method achieves a lower success count on MDSLEN, a higher success count on SLMD, and an equal success count on SLSTR. However, these differences are not significant.

We provide a comparison of the best train error by generations in~\figref{fig:error}. PushGP+EP+ARM holds a much faster convergence speed compared to PushGP; however, it is slightly slower than PushGP+HP+ARM.

In the case of solving the sub-problems and then the composite problems, PushGP+EP+ARM achieves a better performance in train error, success count, and faster convergence, compared to the original PushGP~\cite{helmuth2018program}. However, its performance in both success count and convergence speed is worse than PushGP+HP+ARM without statistical significance.

\section{Experiment II: Sequential KDPS Task}
\label{sec:experiment-ii}
In this second experiment, we test the entire KDPS process in~\figref{fig:kdps}. That is, using PushGP+ARM+EP to solve a sequence of problems. They are the six problems in the last experiment, namely SL, CSL, MD, MDSLEN, SLMD, and SLSTR. Every time a problem is solved, we extract subprograms from its solution and store them in the archive. This archive will be used when solving the next problem.

\subsection{Comparison methods}
\begin{itemize}
\item \textbf{PushGP+EP+ARM}: solving a sequence of PS problems with PushGP+EP+ARM in the way as in~\figref{fig:kdps}; every time a problem is solved, the subprograms are extracted by EP from its solution and added to the archive. This archive is used by PushGP+ARM when solving the next problem.
\item \textbf{PushGP}: solving a sequence of PS problems independently using PushGP~\cite{helmuth2018program}.
\end{itemize}

\subsection{Experimental procedures}
We solve six problems, namely MD, CSL, SL, MDSLEN, SLMD, and SLSTR. The first three problems are from PSB1~\cite{helmuth2015general}. They do not share any sub-problems. The last three problems are the composite problems of MD, CSL, and SL (\secref{sec:experiment-i}). Any pair of the three composite problems share a sub-problem.

For PushGP+EP+ARM, we run a procedure as in~\figref{fig:kdps}. We solve the first problem (MD) with the original PushGP (i.e., PushGP+ARM with an empty archive) and the rest problems with PushGP+ARM. We initialize an empty subprogram archive when solving the first problem. For every problem, we run 25 repetitions and take the best and shortest program. We use EP to extract five subprograms from the best and shortest programs. These subprograms are stored in the archive that we initialized before and used in solving the later problems by PushGP+ARM. For PushGP, we solve the six problems independently in the same order with PushGP+EP+ARM. However, no subprograms are stored and used. 

We provide results of solving the six problems in two different orders. Order 1 solves simple problems at first and later harder ones while Order 2 is a reverse order of Order 1.

\begin{itemize}
\item \textbf{Order 1}: MD\textrightarrow CSL\textrightarrow SL\textrightarrow MDSLEN\textrightarrow SLMD\textrightarrow SLSTR
\item \textbf{Order 2}: SLSTR\textrightarrow SLMD\textrightarrow MDSLEN\textrightarrow SL\textrightarrow CSL\textrightarrow MD
\end{itemize}

We use the same parameter settings as in~\secref{sec:experiment-i}. We present the results and the statistical test in a similar manner as in~\secref{sec:experiment-i}.

\subsection{Experimental results of Order 1}
\begin{figure*}
\centering
\includegraphics[width=\textwidth]{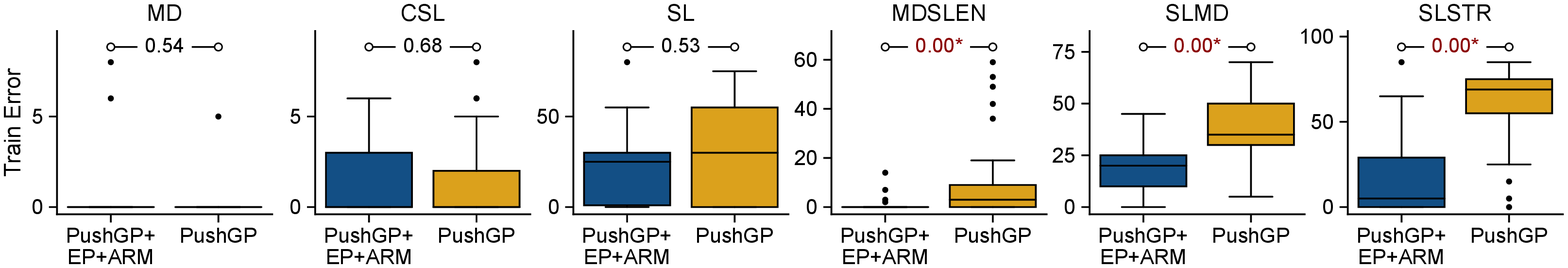}
\caption{\textbf{Experiment II, Order 1}: Train error in 25 runs. The value on a line segment is the p-value of Wilcoxon rank sum test between two groups. The p-value is marked with an asterisk and red color if the difference between two groups is significant. PushGP+EP+ARM holds a significantly lower train error than PushGP on MDSLEN, SLMD, and SLSTR.}
\label{fig:train2}
\end{figure*}
\begin{figure*}
\centering
\includegraphics[width=\textwidth]{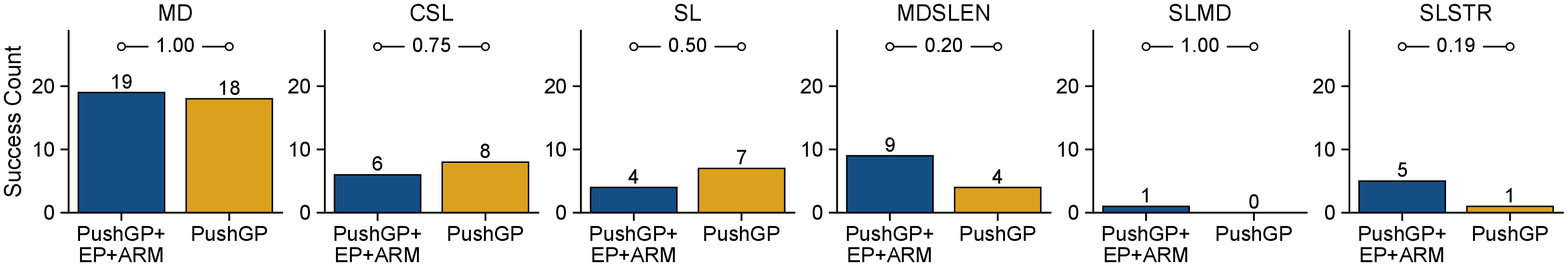}
\caption{\textbf{Experiment II, Order 1}: Test success count in 25 runs. The number above a bar is the success count. The value on a line segment is the p-value of Fisher's exact test between two groups. The p-value is marked with an asterisk and red color if the difference between two groups is significant. PushGP+EP+ARM holds a higher success count than PushGP on MD, MDSLEN, SLMD, and SLSTR; however, without statistical significance.}
\label{fig:count2}
\end{figure*}
\begin{figure*}
\centering
\includegraphics[width=\textwidth]{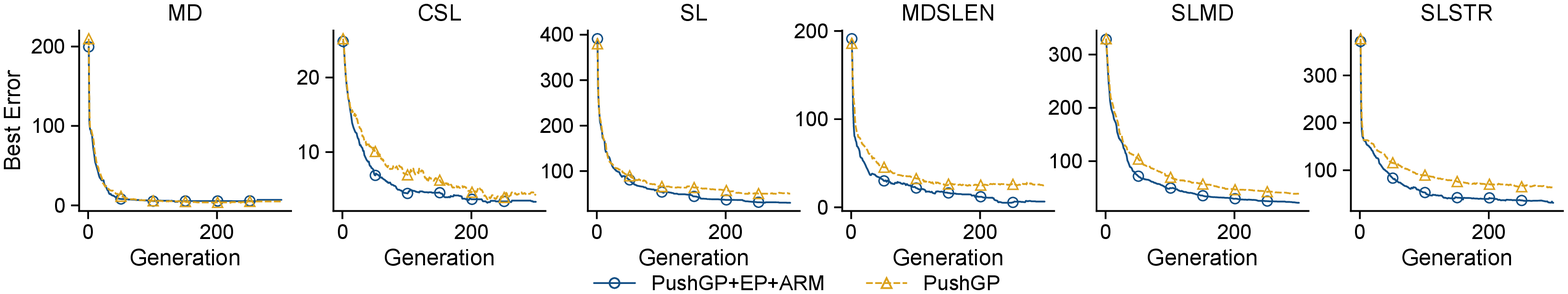}
\caption{\textbf{Experiment II, Order 1}: Average of the best train error in the population by generations. PushGP+EP+ARM converges much faster than PushGP on the five problems except MD.}
\label{fig:error2}
\end{figure*}

In~\figref{fig:train2}, PushGP+EP+ARM holds a significantly lower train error than PushGP on the three composite problems, while the difference on the rest three problems is not significant. \figref{fig:count2} shows the test success count of PushGP+EP+ARM and PushGP. The test success count of PushGP+EP+ARM is lower than PushGP on CSL and SL but higher than PushGP on MSDLEN, SLMD, and SLSTR. All these difference is not statistically significant through Fisher's exact test. On MD, the difference between the two methods is very small since it is solved by two equivalent methods. \figref{fig:error2} provides the best error in the population by generations of both methods. It is obvious that PushGP+EP+ARM converges faster than PushGP on all problems except MD.

\subsection{Experimental results of Order 2}
\begin{figure*}
\centering
\includegraphics[width=\textwidth]{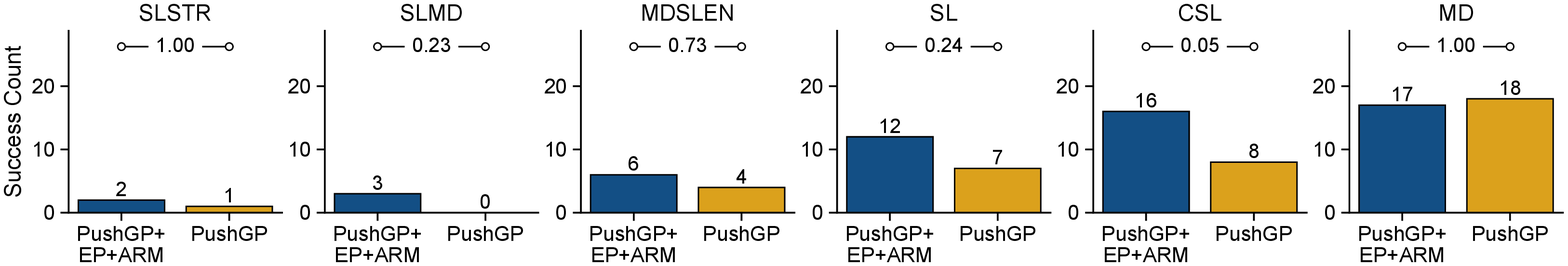}
\caption{\textbf{Experiment II, Order 2}: Test success count in 25 runs. The number above a bar is the success count. The value on a line segment is the p-value of Fisher's exact test between two groups. The p-value is marked with an asterisk and red color if the difference between two groups is significant. PushGP+EP+ARM holds a higher success count than PushGP on SLSTR, SLMD, MDSLEN, SL, and CSL; however, without statistical significance.}
\label{fig:count3}
\end{figure*}

In~\figref{fig:train3}, PushGP+EP+ARM holds a lower train error than PushGP on all problems without statistical significance. \figref{fig:count3} shows the test success count of PushGP+EP+ARM and PushGP. The test success count of PushGP+EP+ARM is higher than PushGP on on most of the problems except MD. Especially on CSL, PushGP+EP+ARM gets 16 success while PushGP only gets 8. However, all these difference is not statistically significant through Fisher's exact test. \figref{fig:error3} provides the best error in the population by generations of both methods. It is obvious that PushGP+EP+ARM converges faster than PushGP on all problems except SLMD.

Therefore, when solving a sequence of problems, PushGP+EP+ARM achieves a better optimization performance (i.e., train error and convergence speed). This performance finally leads to a higher test success count, however, without statistical significance.

\subsection{Discussion}
We find that the PushGP+EP+ARM in~\secref{sec:experiment-i} is better than the PushGP+EP+ARM in~\secref{sec:experiment-ii} with Order 1, in terms of test success on the three composite problems. Though their algorithms are the same, they have at least two differences.

\begin{enumerate}
\item In~\secref{sec:experiment-ii}, PushGP+EP+ARM adds five subprograms to the archive after solving one problem. Therefore, the size of the archive is 15, 20, and 25 when solving MDSLEN, SLMD, and SLSTR, respectively. However, in~\secref{sec:experiment-i}, PushGP+EP+ARM uses archives with 10 subprograms (five for one sub-problem). A larger subprogram archive makes it harder to select helpful subprograms by the adaptive strategy in~\algref{alg:arm}.
\item CSL and SL are solved in different conditions in the two experiments. In~\secref{sec:experiment-ii}, CSL is solved with subprograms from MD; SL is solved with subprograms from MD and CSL. However, no subprogram is used when solving MD, CSL, and SL in~\secref{sec:experiment-i} (i.e., they are solved by the original PushGP~\cite{helmuth2018program}). Moreover, MD, CSL, and SL do not share the same sub-problems. Therefore, solving CSL and SL with PushGP+EP+ARM is not as good as with the original PushGP (as shown in the first three subfigures of~\figref{fig:count2}). Thus, the subprograms extracted in~\secref{sec:experiment-ii} is not as good as in~\secref{sec:experiment-i}. These subprograms further influence the performance of PushGP+EP+ARM in~\secref{sec:experiment-ii} in solving the later problems MDSLEN, SLMD, and SLSTR. This issue is called ``negative transfer''.
\end{enumerate}

\begin{figure*}
\centering
\includegraphics[width=\textwidth]{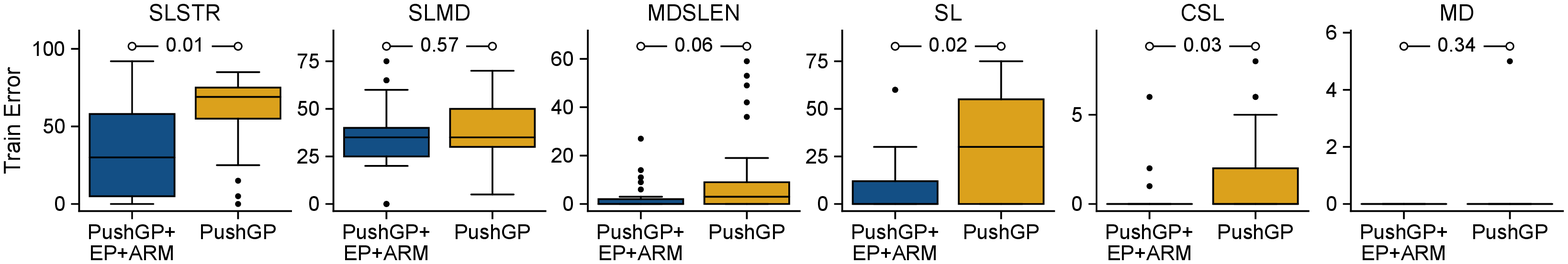}
\caption{\textbf{Experiment II, Order 2}: Train error in 25 runs. The value on a line segment is the p-value of Wilcoxon rank sum test between two groups. The p-value is marked with an asterisk and red color if the difference between two groups is significant. PushGP+EP+ARM holds a lower train error than PushGP on all the problems without statistical significance.}
\label{fig:train3}
\end{figure*}

\begin{figure*}
\centering
\includegraphics[width=\textwidth]{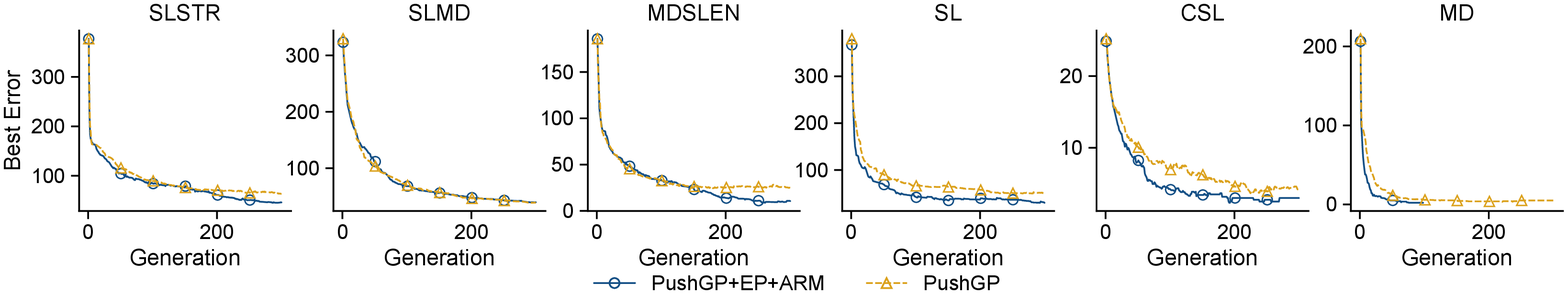}
\caption{\textbf{Experiment II, Order 2}: Average of the best train error in the population by generations. PushGP+EP+ARM converges much faster than PushGP on the five problems except SLMD.}
\label{fig:error3}
\end{figure*}

In~\secref{sec:experiment-ii} (problems in Order 2), we find that PushGP+EP+ARM holds a lower success count on MD. However, on MD, the training error of all runs with PushGP+EP+ARM is 0 (\figref{fig:train3}). Moreover, \figref{fig:error3} shows PushGP+EP+ARM converges much faster than PushGP. This observation may indicate an over-fitting issue with the proposed method.

\section{Conclusions}
\label{sec:conclusions}
In this study, we introduced a problem called Knowledge-Driven Program Synthesis (KDPS) problem. KDPS requires an agent to solve a sequence of related PS problems. To solve KDPS, we proposed a method based on PushGP~\cite{helmuth2018program}. This method consecutively solves programming tasks, extracts subprograms from the solutions as knowledge, and uses these subprograms to solve the next problem. To extract subprograms from the solution of a solved problem, we proposed the Even Partitioning (EP) method; to use these subprograms, we applied Adaptive Replacement Mutation (ARM)~\cite{yifan2022incorporating}.

We compared our proposed method (PushGP+EP+ARM) with the original PushGP~\cite{helmuth2018program} and a method extracting subprograms by humans (PushGP+HP+ARM). Our PushGP+EP+ARM achieved a significantly better train error, success count, and convergence speed than PushGP. The performance of our proposed method is slightly worse than PushGP+HP+ARM. We further compared our PushGP+EP+ARM with the original PushGP in solving a sequence of problems. Our method achieved a better train error and convergence speed. Our PushGP+EP+ARM also holds a higher test success count, however, without statistical significance.

The current method to automatically construct the subprogram archive is rather naive. We would like to improve this method in our future work. In the discussion section in~\secref{sec:experiment-ii}, PushGP+EP+ARM has limitations in dealing with the growing subprogram archive after solving more problems. Moreover, PushGP+EP+ARM also suffers from the ``negative transfer'' of the subprograms from the unrelated problems. A part of our future work is to fix these issues. Methods such as a more efficient adaptation or filtering strategy on subprograms are promising to solve the two limitations. Additionally, the strategy of extracting and using knowledge could be applied to problems other than program synthesis, such as training soft robotics~\cite{pigozzi2022evolving}.


\bibliographystyle{IEEEtran}
\bibliography{ref}

\end{document}